\documentclass[sigconf]{acmart}

\AtBeginDocument{%
  \providecommand\BibTeX{{%
    \normalfont B\kern-0.5em{\scshape i\kern-0.25em b}\kern-0.8em\TeX}}}

\copyrightyear{2021}
\acmYear{2021}
\setcopyright{rightsretained}
\acmConference[SIGIR '21] {Proceedings of the 44th International ACM SIGIR Conference on Research and Development in Information Retrieval}{July 11--15, 2021}{Virtual Event, Canada.}
\acmBooktitle{Proceedings of the 44th International ACM SIGIR Conference on Research and Development in Information Retrieval (SIGIR '21), July 11--15, 2021, Virtual Event, Canada}
\acmPrice{}
\acmISBN{978-1-4503-8037-9/21/07}
\acmDOI{10.1145/3404835.3463114}

\begin{document}
\fancyhead{}




\title{Visual Question Rewriting for Increasing Response Rate}


\author{Jiayi Wei}
\email{jwei37@ucsc.edu}
\affiliation{%
  \institution{University of California, Santa Cruz}
  \country{USA}
  }

\author{Xilian Li}
\email{xli237@ucsc.edu}
\affiliation{%
  \institution{University of California, Santa Cruz}
  \country{USA}
  }

\author{Yi Zhang}
\email{yiz@ucsc.edu}
\affiliation{
 \institution{University of California, Santa Cruz}
 \country{USA}}

\author{Xin Eric Wang}
\email{xwang366@ucsc.edu}
\affiliation{
 \institution{University of California, Santa Cruz}
  \country{USA}
  }

\renewcommand{\shortauthors}{Jiayi Wang, et al.}

\begin{abstract}
  When a human asks questions online, or when a conversational virtual agent asks a human questions, questions triggering emotions or with details might more likely to get responses or answers. we explore how to automatically rewrite natural language questions to improve the response rate form people. In particular, a new task of Visual Question Rewriting (VQR) task is introduced to explore how visual information can be used to improve the new question(s). A data set containing $\sim$4K bland$\&$attractive question-images triples is collected. We developed some baseline sequence to sequence models and more advanced transformer-based models, which take a bland question and a related image as input, and output a rewritten question that's expected to be more attractive. Offline experiments and mechanical Turk based evaluations show that it's possible to rewrite bland questions in a more detailed and attractive way to increase response rate, and images can be helpful.
\end{abstract}



\maketitle

\section{Introduction}
\begin{figure}[h]
    \centering
    \includegraphics[width=\linewidth]{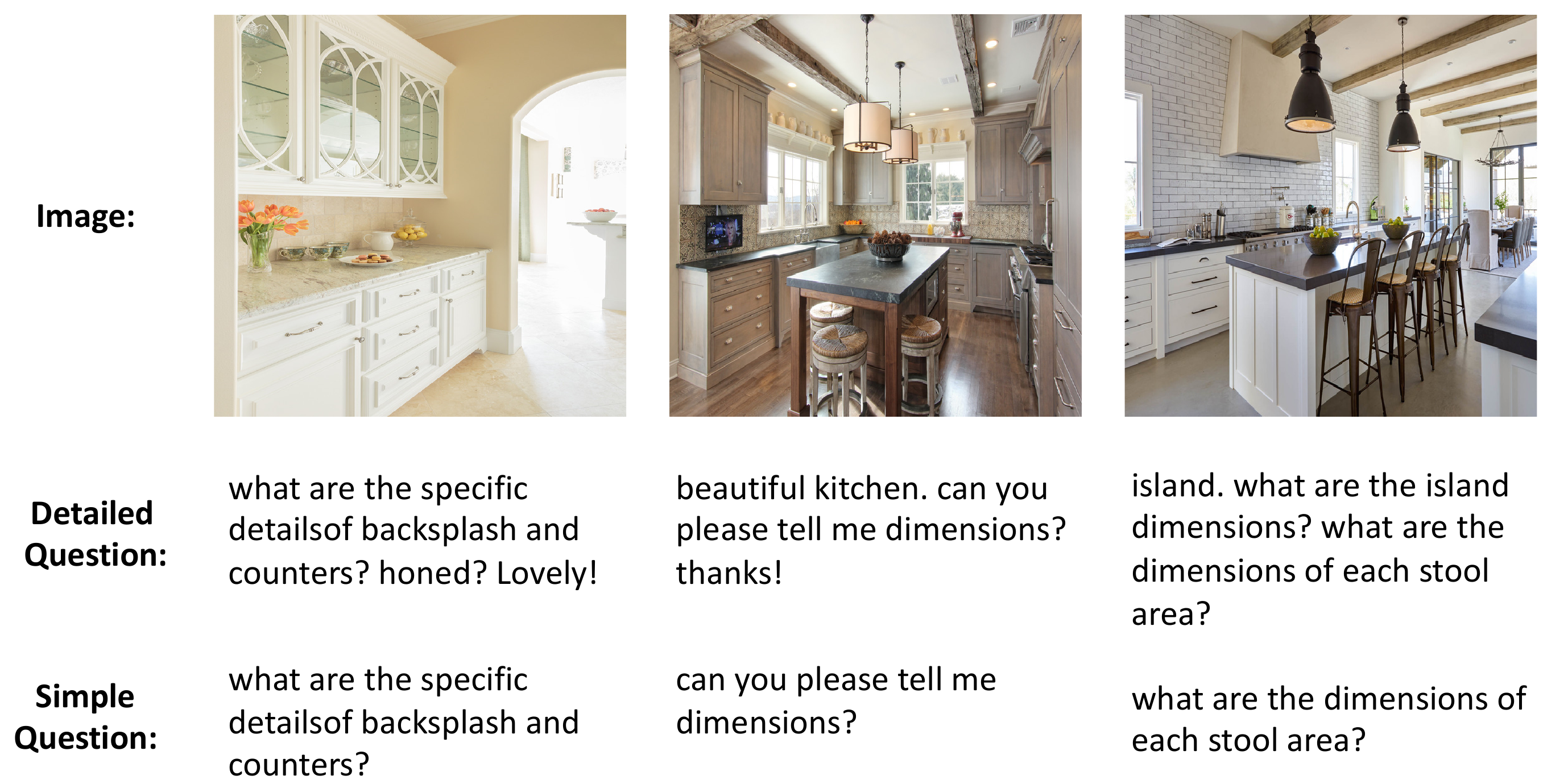}
    \caption{Example of image, simple question and detailed question triples.}
    \Description{Visual question rewriting examples: image, simple question and detailed question triples}
    \label{fig:data_sample}
\end{figure}

When buying complex products such as real estates, furniture, automobiles, buyers often shop locally and ask local store staff various product related questions before making a purchasing decision. However, with the advance of online shopping or virtual reality technologies, and with the impact of covid-19, people are starting to  make complex purchases and asking questions online. Some questions are answered by unpaid specialists, sellers or other buyers, however, many questions are not answered.  \\
After analyzing online Q\&A web sites or discussion forums, we found more attractive questions, such as those with details or emotional words, are more likely to be answered. However, majority normal users are not good at writing attractive questions, or may not take the effort to do so, when shopping around and browsing product images. Thus we propose to automatically rewrite a simple question into a more attractive one, and we expect the visual image associated with the original question might help. \\
This paper proposes a new Visual Question Rewriting (VQR) problem that takes an image and a simple natural language question related to the image as input, and output a more attractive question. We created a data set containing many <image, question, more attractive question> triplets based on Houzz.com, an online community with many questions about published home designs. Some questions on Houzz received answers from the designer who shared the design image. Some examples are shown in Fig \ref{fig:data_sample}. A baseline Seq2Seq \cite{sutskever2014sequence} model and a transformer-based model are implemented to explore whether the proposed research direction is promising and what results look like.


\section{RELATED WORK}
This work is mainly related to and motivated by a large amount of prior research on natural language processing, computer vision and the interplay between the two areas.

\textbf{Natural Language Processing} has been studied since the late 1940s. NLP field has several important sub-tasks. Summarizing, text generation, story writing, text simplification, rewriting questions in context of a dialog systems, and grammar error correction are existing NLP tasks related our problem. The performances of those tasks have improved significantly in the last several years due to the advance of deep learning technologies and the availability of large scale training data. The models we proposed in this paper are motivated by these prior research work, especially the recent advances \cite{radford2018improving,goldfarb2019plan,xu2019alter,elgohary2019can,radford2019language}. However, the goal of our new rewriting task is different: we are focusing on extending and enriching  a user's input so that it's more attractive. 


\textbf{Computer vision(CV)} systems have improved significantly over the last decade and perform better than human in some standard classification or object detection tasks \cite{simonyan2014very,he2016deep}. Over the last several years, two main tasks combining natural language and computer vision have emerged: Visual Question Answering \cite{antol2015vqa,goyal2017making,anderson2018bottom} and Image Captioning \cite{lin2014microsoft}. Existing VQA tasks mainly  focus on selecting the best answer from candidates or produce a short answer for questions based on given images, while our new task focuses on how to generate a more attractive question to increase response rate, based on visual information and the original text question. Usually, the answer of our questions can not be derived from the images and needs to be answered by a human. These differences make our problem novel and more difficult to evaluate. That's why we start with a specific domain in this paper.\\

\section{Dataset Collection}
To support the new research, we crawled images and their related questions from Houzz.com. $\sim$3K images with $\sim$18K questions are collected. To reconstruct a data set with <image, attractive question, bland question> triplets,  questions that have some response(s) and contain more than 1 sentence are used, assuming those questions are attractive. For each attractive question, the key question sentence is selected as the corresponding bland question. As a result, the new VQR dataset contains $\sim$4K image-attractive question-bland question triples. We split it into $4:1$ for training and testing. 

\section{Methods}

As a preliminary research, we evaluated 4 different straightforward model architectures. These models are motivated by existing research in NLP\cite{wolf2020transformers} and image captioning\cite{fang2015captions,chen2015mind}. They either take the original question or (original question, associated image) as input, and output a new question. 

\subsection{Baseline Models}

\begin{figure}[h]
    \centering
    \includegraphics[width=\linewidth]{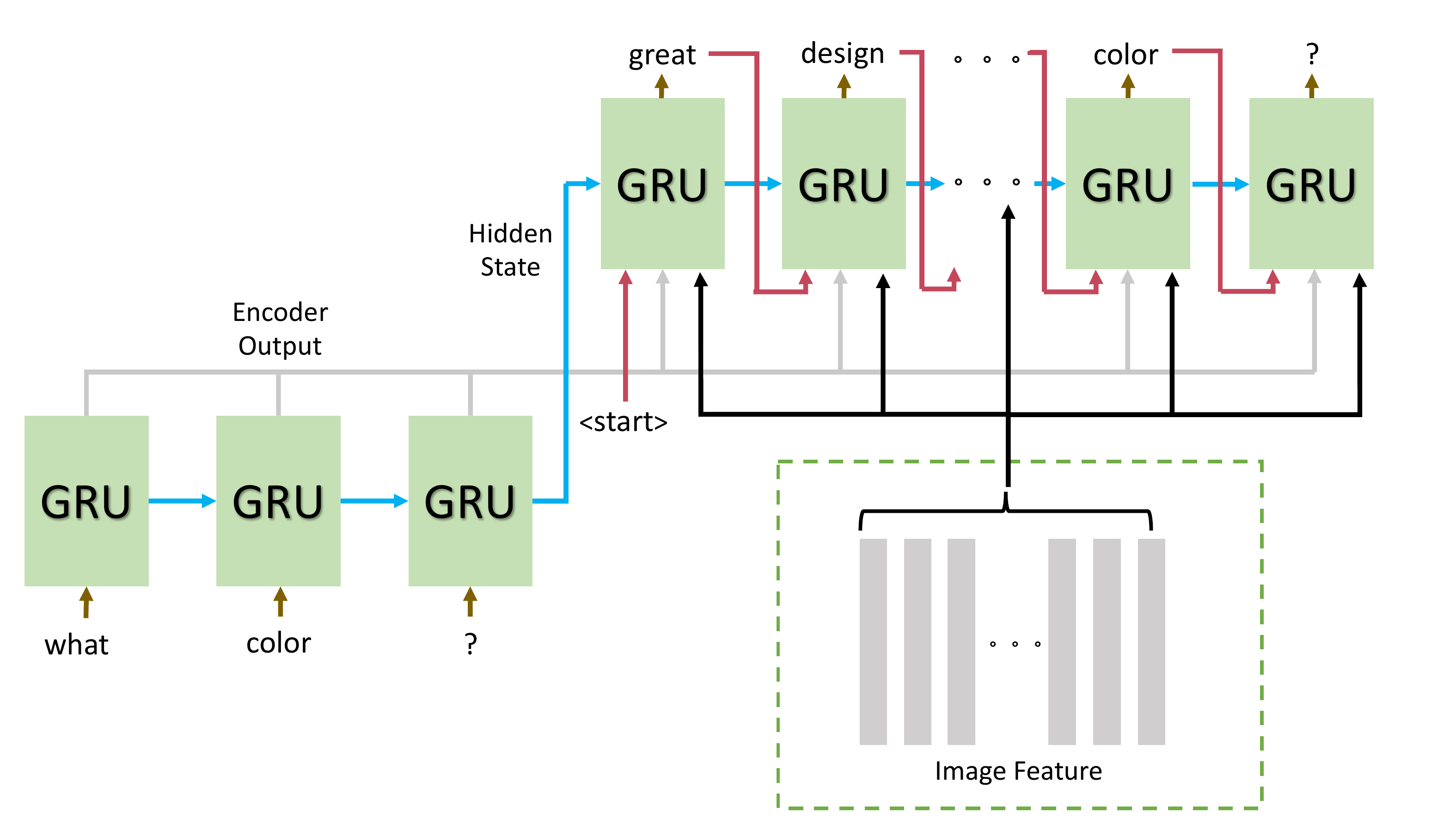}
    \caption{Baseline model illustration. The architecture without the green dashed box is the pure language model. The part in the green dashed box is for the visual part.}
    \label{fig:s2s}
\end{figure}
We implemented a basic Seq2Seq architecture for natural language generation as the baseline solution. In order to explore the impact of the image input features, we did experiments with or without the visual input, as illustrated in Fig \ref{fig:s2s}.\\
Firstly tokenizing all text information (input and output questions), provided us a vocabulary of $\sim$5K words. Then all input and targeting sentences were mapped into sequences with length of 30 and 50, respectively. The baseline model has 2 layers of Gated recurrent units(GRUs) \cite{cho2014learning} for both encoder and decoder. With the output from the encoder, the attention mechanism was applied to generate the attention weights which are used to pass weighted sum of encoder output to the decoder part. The attention equations \cite{luong2015effective} used in our work are as below, where $h_{s}$ denotes encoder output and $h_{t}$ denotes the current hidden state vector, $v,W$ are model parameters, $\alpha_{ts}$ are attention weights, $c_{t}$ is a context vector for the decoder:

\begin{equation}
    score(h_{t},h_{s}) = v_{\alpha}^{T}tanh(W_{\alpha}[h_{t};h_{s}])
\end{equation}

\begin{equation}
    \alpha_{ts} = softmax(score(h_{t},h_{s}))
\end{equation}

\begin{equation}
    c_{t} = \sum_{s}(\alpha_{ts} h_{s})
\end{equation}

Then, the embedding vector from the previous position and the acquired $vector$ are concatenated together and passed into the decoder. In the experiments, we set the length of the embedding vector and hidden status vector $256$ and $768$ respectively.

Next, visual information is added to the model to see whether it can boost the performance. All images were resized as $224*224$ and feed to a pre-trained ResNet-50 \cite{he2016deep}. The feature map before the fully-connected layer was used as the input for our model (Fig \ref{fig:s2s}). After the forward pass, we received a $7*7*2048$ feature set. We flattened the feature map to $49*2048$. We implemented the similar attention method on the flattened image feature (Fig \ref{fig:s2s}), because the $7*7$ grid possess the position information on this patch. The language generation part remained the same here.

\subsection{Transformer Based Model}

The self-attention \cite{vaswani2017attention} model is originally designed for natural language understanding (NLU) tasks. BERT \cite{vaswani2017attention,devlin2018bert,dong2019unified} drives it to a higher level, which updates the state-of-the-art metrics in most NLU tasks. Transformer-based models allow the encoder and decoder to see the entire input sequence all at once and to directly model these dependencies using attentions. More conveniently, BERT model pre-trained on a large corpus is publicly available, which enables researchers to start with a good model and fine-tune on their specific tasks. \\

Much work has been attempted to combine BERT with Seq2Seq models. UniLM \cite{dong2019unified} is one of the most successful and decent one. The original BERT model sets mask on words randomly to force the model better understand the generic context. Similarly, UniLM treats Seq2Seq as language completion. In this way, the attention on the input and output are bi-directional and uni-directional respectively, which can satisfy the requirement for language generation. Besides, with this specific mask, the language generation task is completed with a single BERT model and powerful pre-trained BERT weights. \\

\begin{figure}[h]
    \includegraphics[width=\linewidth]{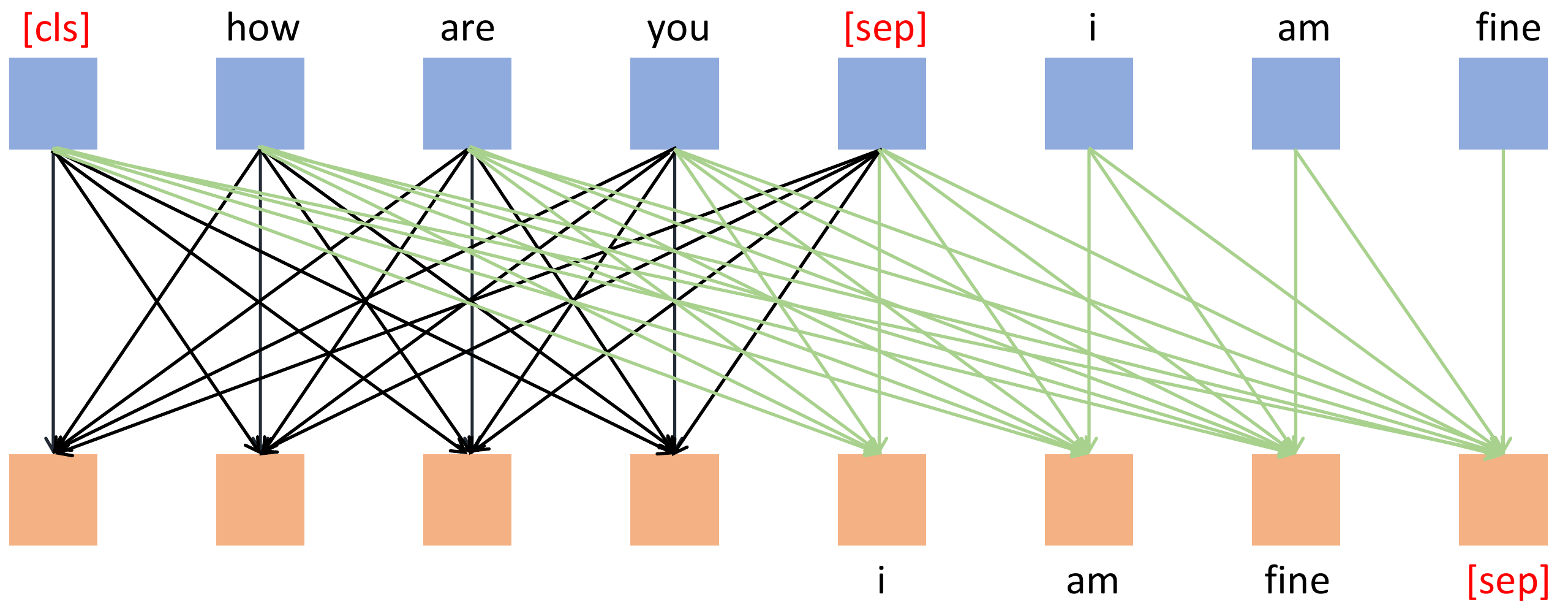}
    \caption{Illustration of attention mask in the transformer based language generation model.}
    \label{fig:unilm}
\end{figure}

\begin{table}
    \caption{Model performance on our VQR data.}
    \label{tab:result}
    \begin{tabular}{ccccl}
      \toprule
      & baseline & baseline$+$vis & transformer & transformer$+$vis  \\
      \midrule
      BLEU & 0.194 & Nan & 0.528 & 0.547 \\
      RG-1 & 0.198 & Nan & 0.753 & 0.765 \\
      RG-2 & 0.167 & Nan & 0.614 & 0.651 \\
      RG-L & 0.307 & Nan & 0.870 & 0.885 \\
      \bottomrule
    \end{tabular}
\end{table}

\begin{figure}[h]
    \centering
    \includegraphics[width=\linewidth]{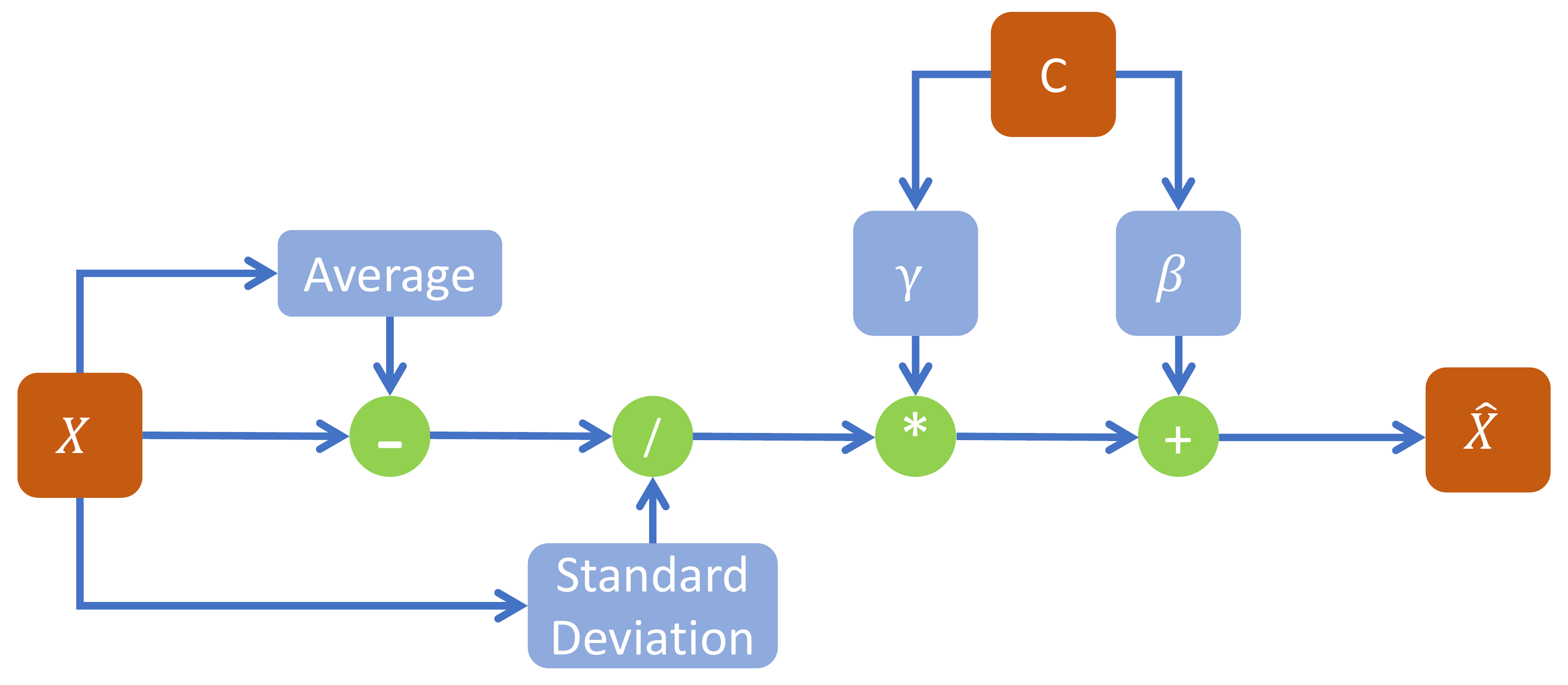}
    \caption{The illustration of conditional layer normalization.}
    \label{fig:cn}
\end{figure}

We firstly evaluated the transformer-based language model without image input in Figure \ref{fig:unilm}. Then, we did experiments by adding image-based inputs, again using ResNet-50 as the feature extractor. We found the attention mechanism on image doesn't work on the baseline method, thus we used the generic image feature instead of the grid image feature, which means the feature after the dense layer is used as our visual input to be combined with the language feature. Visual information is included through conditional normalization \cite{devries2017modulating} (Figure \ref{fig:cn}): 

\begin{equation}
\hat{x} = \gamma(c) \times \frac{x-\mu}{\sigma+\epsilon}+\beta(c)
\end{equation}

\begin{displaymath}
\mu = \frac{1}{N}\sum x, \;\; \sigma^2=\frac{1}{N}\sum (x-\mu)^2
\end{displaymath}

where  $x$ and $c$ are language feature and conditional input (image feature) respectively, $\mu$ and $\sigma^2$ are mean and variance, and $\epsilon$ is a small constant number. This conditional normalization are firstly proposed and widely utilized as conditional batch normalization \cite{devries2017modulating, huang2017arbitrary, miyato2018cgans}. In our work, we modified the pre-trained BERT to generate language sequence and applied layer normalization as the main normalization method. $\beta$ and $\gamma$ variables in BERT's normalization layers are changed to control the sequence output. In traditional normalization layers, $\beta$ and $\gamma$ are learned through parameter updates. However, with conditional normalization, $\beta$ and $\gamma$ are generated from input, thus the normalization layer output is conditional on its side input.  Because $\beta$ and $\gamma$ in the pre-trained model have fixed length, a dense layer is used to convert the image features into the required length. 

\section{Experiments and Analysis}

This section presents the experiment details and results. The RNN based Seq2Seq models are called baseline and baseline$+$vis, and the transformer based models are called transformer and transformer$+$vis.


\begin{figure*}
    \centering
    \includegraphics[width=\linewidth]{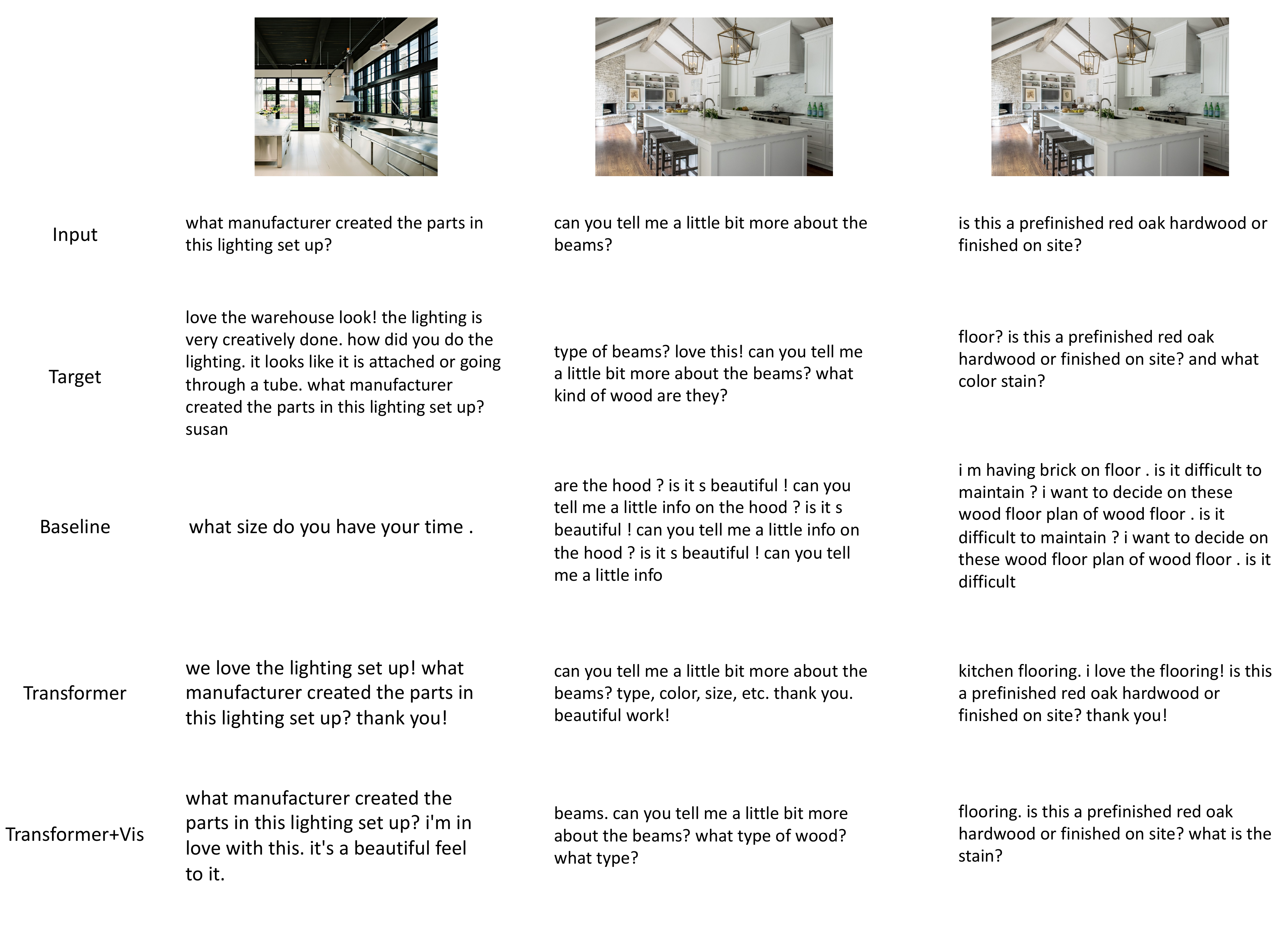}
    \caption{Examples of result on the VQR test data.}
    \label{fig:resultexample}
\end{figure*}
The results are shown in Table \ref{tab:result}. The baseline models performs poorly compared with the transformer models. The baseline$+$vis model didn't converge at the training stage, probably due to the lack of pre-trained models and the number of training data in our task is not enough for leaning embedding. 
Transformer models perform much better, probably because BERT weights used are pre-trained on large amount of training data sets across multiple language tasks jointly, which makes BERT more general and not over-fit on any single task. Besides, with attention mask specified for language generation, the attention of the input part is bidirectional and the attention of the output part is unidirectional. Therefore, the transformer model performs better than the baseline model, which is common for other NLP tasks.\\
With the help of conditional layer normalization, the related image features really contribute to language generation and improve performance. Rather than the bigger feature map, the image feature vector before $softmax$ contains more generic context information. Besides, using the generic context as input, there is no need to learn attentions based on VQR data, which seems to be a difficult task and the main reason for baseline$+$vis not converging. Adding image feature through layer normalization effectively controls its influence, especially with zero initialization, as this accelerates the training process and avoids the harm of warming up.


Further analysis was done by examining some representative examples. As shown in Figure \ref{fig:resultexample}, the baseline model does not know when and where to stop, a common problem shared by many standard RNN methods. It also fails to acquire the most obvious pattern in the training data: copy the original question. As we mentioned before, the baseline$+$vis model doesn't converge during training, thus produce weird results, so we won't show it here. The output of the transformer models seems more attractive, with emotional parts added at the beginning or end of the original bland sentence. High level details "type, color, size, etc." could be added occasionally (middle example) by the transformer model, this is due to the trained language model, instead of the influence of the associated image. On the other hand, the transformer$+$vis model is able to add new information inferred from the image, such as "what type of wood?", "what is the stain". This suggests image helps adding extra details to the bland question.


\section{Human Evaluations}

We hired several Amazon Mechanical Turks to evaluate the generated results on test data. 
For each image, we present the question generated by transformer$+$vis model and the original bland question, let 3 different workers choose which question is more attractive independently, then use the majority vote as the final judge for each image. 
Out of 767 images, 652 rewriting questions ($85.8\%$) are judged to be better than simple bland questions. This demonstrates that the rewriting model successfully generates detailed and emotional questions and makes the question more attractive. 

We also hired Turks to compare questions generated by transfer vs transformer$+visual$. 410 questions ($53.4\%$) generated by the transformer$+$vis model are preferred by workers, which suggest extra information brought by the image input might make the rewriting more attractive. 



\section{CONCLUSIONS}
We propose a new task of Visual Question Rewriting (VQR) to convert a natural language question into a more attractive one, so that it's more likely to be answered. A new data set was collected and will be shared with the research community. As a first step, we implemented two straightforward approaches: a standard sequence to sequence model and a transformer-based model, and found the transformer based model performs much better. We modified the original BERT model, and transferred image features as conditional layer normalization input to influence the output sentences, and found this transformer$+$visual model can rewrite a question with more details. User evaluation demonstrated visual question rewriting is more attractive, thus a promising direction to increase question response rate.

Although conversational IR is not the focus of this paper, how to make conversational agents, such as conversational recommendation or QA systems more attractive, is a major challenge in practice. This work can also be adapted to re-write questions made by a virtual assistant to increase user engagements with the system.



\bibliographystyle{ACM-Reference-Format}
\bibliography{sample-base}


\end{document}